\documentclass[10pt,twocolumn,letterpaper]{article}

\usepackage{wacv}              

\usepackage{graphicx}
\usepackage{amsmath}
\usepackage{amssymb}
\usepackage{booktabs}
\usepackage{comment}
\usepackage{float}
\usepackage{multirow}

\usepackage[pagebackref,breaklinks,colorlinks]{hyperref}

\usepackage[capitalize]{cleveref}
\crefname{section}{Sec.}{Secs.}
\Crefname{section}{Section}{Sections}
\Crefname{table}{Table}{Tables}
\crefname{table}{Tab.}{Tabs.}

\def\wacvPaperID{2009} 
\def\confName{WACV}
\def\confYear{2024}

\begin{document}

\title{The Background Also Matters: Background-Aware Motion-Guided Objects Discovery  

}

\author{Sandra Kara
\and
Hejer Ammar
\and
Florian Chabot
\and
Quoc-Cuong Pham\\
\and
Universit\'e Paris-Saclay, CEA, List, F-91120, Palaiseau, France\\
{\tt\small \string{firstname.lastname\string}@cea.fr}}
\maketitle

\begin{abstract}
Recent works have shown that objects discovery can largely benefit from the inherent motion information in video data. However, these methods lack a proper background processing, resulting in an over-segmentation of the non-object regions into random segments. This is a critical limitation given the unsupervised setting, where object segments and noise are not distinguishable. To address this limitation we propose BMOD, a Background-aware Motion-guided Objects Discovery method. Concretely, we leverage masks of moving objects extracted from optical flow and design a learning mechanism to extend them to the true foreground composed of both moving and static objects. The background, a complementary concept of the learned foreground class, is then isolated in the object discovery process. This enables a joint learning of the objects discovery task and the object/non-object separation. The conducted experiments on synthetic and real-world datasets show that integrating our background handling with various cutting-edge methods brings each time a considerable improvement. Specifically, we improve the objects discovery performance with a large margin, while establishing a strong baseline for object/non-object separation. 
\end{abstract}

\section{Introduction}
\label{sec:intro}

Deep learning-based approaches have demonstrated significant success in addressing a wide array of computer vision tasks \cite{DL}. However, the high performance of these methods heavily relies on the availability of abundant labeled data: sparse labels or compromised label quality impairs the effectiveness of supervised approaches \cite{labels}. This limitation becomes challenging when tackling dense tasks like segmentation, where the acquisition of accurate labels requires considerable resources. This observation has motivated numerous studies to propose alternative architectures, including weakly supervised \cite{IAM, maskconsist}, semi-supervised \cite{semi_seg, semi_inst_seg}, and unsupervised methods \cite{LOST, tokencut}, aiming to tackle vision tasks with minimal supervision.\newline

\noindent In this work, we address the task of localizing objects in videos without the use of human annotations. This task, which is commonly approached as a segmentation problem \cite{SAVI, SAVI++, MOTOK}, is particularly suited for video data due to its inherent advantages over static images. Specifically, the motion information derived from videos offers a means to obtain \textit{free} pseudo-labels for moving objects localization. This makes the motivation even stronger to explore self-supervised methods, capable of leveraging motion cues. 
Moreover, the ambiguity surrounding the definition of objects, which remains a challenge for object discovery in images\cite{rOSD,UMOD}, can be addressed in the video data. Specifically, relying on motion cues to localize objects provides, by design, a definition of what an object is: we consider as object any entity that could exhibit an independent motion. This definition is even in line with human perception as demonstrated in \cite{SPELKE}: in our perception, we divide the observed scene into parts that are capable of moving while remaining connected. Some recent approaches draw inspiration from and build upon this result, to address moving objects localization \cite{segSPELKE}.\newline

\noindent Recently, object-centric learning architectures have demonstrated a significant potential for solving the object discovery task \cite{SA}. It emerges as a new deep learning based approach to decompose the input image into \textit{meaningful} regions, in an unsupervised way. Although initially validated on simple synthetic image datasets, many subsequent works have proposed variants of this architecture to scale it to video data as well as to more complex scenarios. Those primarily concentrate on modifying the reconstruction space (optical flow \cite{SAVI}, depth \cite{SAVI++}) and enhancing the encoder's capability \cite{STEVE}. More recently, some works introduced the use of motion cues to direct the learning process of slots and provided evidence of the effectiveness of this guidance signal in solving objects discovery in complex scenarios \cite{DOM, MOTOK}. \newline

\begin{figure}[t]
  \centerline{\includegraphics[width=8.5cm]{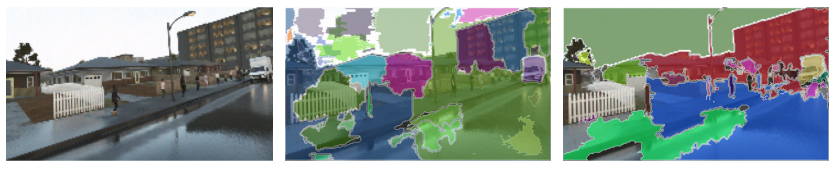}}
  \caption{\textbf{Illustration of the addressed problem.} Results from \cite{DOM} showing background over-segmentation in both settings: unsupervised (middle) and using motion supervision (right). When the ground truth is not available, foreground objects cannot be automatically separated from the background.}
  \label{fig:demo_intro}
\end{figure}

\noindent Our work fits into this same line of research, but tackles a specific problem that is not covered by existing methods, namely the background control in the object discovery task. Previous works did not focus on learning the background pattern, which results in the background being split across the slots into \textit{noise} regions, as illustrated in figure \ref{fig:demo_intro}. This over-segmentation of the scene is not even penalized by the commonly used metrics, since the segmentation quality is evaluated on foreground regions only. However, in a real-world setting where ground truth is not available, it is impossible to distinguish between objects and background segments. The aim of this work is therefore to learn this object/non-object boundary, while solving the multiple objects discovery task. \\

\noindent We propose to leverage motion cues to jointly learn the multiple objects discovery and the objectness task (foreground/background separation). The motion cues are moving objects masks, extracted from optical flow. For the first task, each motion mask is used to guide one slot's attention. For the second, we propose to learn the generalization from the moving foreground (summed motion masks) to the \textit{true} foreground containing both moving and static objects. The complementary mask, which is the background, is positioned within a specific slot, competing with all others, to isolate its distinct pattern.\newline

\noindent Our contributions can be summarized as following:
\begin{itemize}
    \item We propose BMOD (Background-aware Motion-guided Objects Discovery), a simple yet effective learning mechanism for modeling the background while solving the object discovery task. To the best of our knowledge, this is the first method that addresses these two tasks concurrently, without the need for human supervision. 
    \item We demonstrate that modeling the background not only allows for a more precise objects discovery (automatic filtering of noise segments), but also improves the localization of foreground objects. This validates our insight that controlling the background reduces the amount of noise captured by the slots, making it easier for the model to learn the \textit{object pattern}.
    \item We establish a new baseline for the objectness learning in the object discovery task. For the first time, we introduce the computation of suitable metrics for evaluating the objectness learning task (Jaccard score), or by evaluating the two tasks together (all-ARI).
    \item We demonstrate through comprehensive experiments the effectiveness of our method on the challenging TRI-PD dataset as well as the real-world dataset KITTI. The experiments show that multiple cutting-edge methods derive significant advantage from our objectness learning mechanism, without increasing architectural complexity. Moreover, we show that our method, when enhanced with rich features from the recent DINOv2\cite{dinov2} pretraining, brings about a considerable performance leap. This provides evidence of the representation bottleneck in current methods, which is overcome through the use of improved features.
\end{itemize}

\section{Related work}
\subsection{Objects discovery in images}
Object discovery in images is the task of localizing objects without the use of human annotations. The inception of this task was marked by heuristic-based object proposal methods, which relied on an over-segmentation of the image and various similarity measures to merge \textit{similar} regions hierarchically \cite{SeSe, EB, MCG}. Due to their very low precision, utilizing these object candidates in an unsupervised setting has been challenging.

In the era of deep learning, object discovery has profited from deep features, either derived from CNNs learned through the ImageNet classification task \cite{OSD, rOSD, LOD}, or from the more recent self-supervised pretraining, in particular of vision transformers (ViTs) \cite{tokencut, LOST, UMOD}. In the first category, methods typically aim at discovering the dataset-structure, with the most connected/similar object proposals becoming the top object candidates. In the second category, methods are mostly motivated to investigate unsupervised clustering in the space of self-supervised features, given the segmentation properties exhibited by ViTs \cite{DINO}. In both categories, the methods solely rely on the semantic information learned within the image modality, which limits their ability to separate object instances.

 A recent group of methodologies, known as compositional generative models, has emerged as a deep learning-based alternative to the classical clustering methods \cite{kmeans}. Notably, MONet \cite{MONET} employs an attention mechanism to focus on individual scene parts. Both the input image and the attention map are then passed into a variational auto-encoder module, to only reconstruct the highlighted scene part/object in the corresponding mask. IODINE \cite{IODINE} replaced the one-pass attention mechanism with an iterative inference to refine the understanding of the image over multiple steps. In this same category, SCALOR \cite{SCALOR} adapted the generative process to a larger number of objects, while Slot-Attention \cite{SA} proposed a more efficient object discovery architecture with a single image encoding step. \cite{SA} discovers objects by enforcing the disentanglement within the latent space of an auto-encoder architecture.

All previous methods, whether based on heuristics or deep learning, suffer from the ambiguity of object definition. This limitation prevents both the design of a definition-based method and the establishment of objective evaluation criteria.
Efforts are now being directed towards video data, which provide the means for a more generalized object definition (see section \ref{VOD}). In this work, although we focus on the analysis of video data, we provide comparisons with the latest image-based methods, applied to individual frames. 

\subsection{Objects discovery in videos} \label{VOD}
In this work, we address the problem of discovering objects in videos, which is a distinct task from the video object segmentation (VOS). The latter is more about motion segmentation, with as objective to localize a salient moving object within a video\cite{VOS}. The task we address, in contrast, consists in localizing objects that are capable of moving, even when they remain static in the analyzed sequence.

Object discovery in videos emerges as a promising research area, largely driven by the inherent motion information in videos, compared to static images. The motivation to exploit video data for localizing objects is not new; the earliest methods typically selected regions of interest from object candidates as spatio-temporal tubes, maximizing similarity across videos while maintaining temporal consistency\cite{EURLIER_VOD}.

The recent advent of the slot-attention architecture\cite{SA}, recognised as a promising solution for object discovery, has motivated many efforts to scale it to video data. SAVI \cite{SAVI} and Karazija et al.\cite{karazija} incorporated optical flow as a more task-appropriate reconstruction space for the targeted segmentation task. \cite{SAVI} also proposed the use of weak supervision on the initial frame, such as the centers of objects to be tracked throughout the sequence. By design, \cite{SAVI} presents the limitation of localizing moving objects only.
SAVI++ \cite{SAVI++}, in contrast, also localizes static objects through the reconstruction of the more generic depth signal. \cite{SAVI++} also demonstrated the potential of data augmentations, often under-explored in unsupervised settings. Among methods that utilize motion cues for object discovery, Bao et al. \cite{DOM} introduced an explicit guidance for slots learning, using moving-object masks derived from optical flow. STEVE \cite{STEVE} a concurrent work, investigated the use of a more powerful transformer decoder.
Building upon the findings of these preceding methods, MoTok \cite{MOTOK} proposed a more powerful motion-guided slot attention architecture through a tokenized reconstruction space.

In the previous methods, attention was only allowed to the discovery of foreground objects, without considering a proper background modeling. Our insight, however, is that a proper background modeling prevents the presence of noise regions captured by each slot, which favors the learning of the object structure. We propose in this work a complete motion-guided object discovery architecture to jointly learn the multiple objects localization task and the foreground/background separation.

\subsection{Unsupervised background segmentation}
 Early attempts to solve the task of foreground/background separation focused on the image modality. In simple scenarios with a \textit{neat} background, methods typically relied on thresholding or binary clustering in the color space or other hand-crafted features\cite{colorSpace}.
 
 A more recent category, known as saliency detection methods, aim to extract a salient foreground from the background in an unsupervised way. In particular, LOST and TokenCut \cite{LOST, tokencut} used deep features from pre-trained vision transformers (ViTs)\cite{DINO}. LOST defined object regions as the patches least correlated with the whole image, while TokenCut investigated applying spectral clustering\cite{ncut} to self-supervised ViTs features. More recently, FOUND\cite{found} proposed to discover the background as the class containing the least activated patch in ViT activation maps, and then refine it using a lightweight segmentation head. This method was presented as a way of overcoming the ambiguity of object definition. However, we believe that the problem of object definition remains valid for the background class, since the two are complementary semantic concepts.
 
 In the video modality, the binary segmentation that received significant attention was motion segmentation. Active benchmarks on this subject have been established under the terminology of Video Object Segmentation (VOS)\cite{VOS}. While this provides a well-defined criterion for object identification (i.e. moving objects), we believe this definition is restrictive. Indeed, without also localizing the static objects, we can only achieve a limited understanding of the scene.

Our method, in contrast, proposes by design a foreground/background separation, where the targeted foreground is composed of both moving and static objects. The robustness of our method in complex scenarios is ensured by the use of motion cues, extracted from optical flow, which is typically insensitive to an increasing background complexity in the color space. Moreover, we also decompose the foreground class into object instances, which is not covered by the previous background segmentation methods.

\section{Method}
\subsection{Context: motion guided slot-attention for objects discovery} \label{OCL}

 Since our method is based on a slot attention architecture\cite{SA} involving the use of motion information\cite{DOM}, we first briefly describe these two approaches below.

The slot attention architecture \cite{SA} has been proposed as a deep learning-based alternative for the classical unsupervised clustering methods \cite{kmeans}. 
It consists of an auto-encoder architecture with a latent space that is partitioned into embedding vectors called slots.
The architecture competes among these slots to provide a comprehensive explanation for the input image. 
The mechanism for partitioning the image is encouraged by the use of a small decoder: each slot is individually passed through the decoder. The small decoder being unable to explain the whole scene from one slot, this compels the features to be split across the slots, encouraging image partitioning into \textit{meaningful} regions.

In our method, we build upon a recent variant of slot attention that exploits motion cues to guide slots learning \cite{DOM}. Concretely, the method receives as input a sequence of $T$ video frames. Each frame is passed through a CNN encoder for features extraction. Features from the $T$ frames are then combined using a convGRU module to get spatio-temporal information $H^{t}$, for each frame $I^{t}.$ This representation is then assigned to $K$ slots through the attention module. Specifically, given $k,q,v$ three learnable linear projections, attentions between features $H$ and slots $S$ are computed as $W = \frac{1}{\sqrt{D}} k(H)\cdot q(S) \in \mathbb{R}^{N\times K}$, where $N$ is the feature maps size and $D$ the dimension of features after projection. Attentions are used to update the current slot state $S^{t} = W^{t}v(H^{t})$, where $W^{t}$ are computed using the slot state at frame $I^{t-1}$. \cite{DOM} introduced the use of motion guidance by assuming access to $M$ motion masks for the sequence of $T$ frames. The masks are resized to match the dimensions of the attention maps $W$ and subsequently paired with them via a bipartite matching algorithm. Motion supervision then occurs between these pairs of masks $m$ and the learned attention maps $W$. The method shows that introducing motion cues replaces the initial inductive bias about individual slot decoding, which reduces memory demands. Although this architecture showed generalization ability to objects without corresponding masks, it still suffers from the object/non-object ambiguity, since slots with no supervision may contain either objects or background regions. This motivates our work which we describe in next sections.

\subsection{Modeling the background class using motion cues} \label{ours}

\begin{figure*}[ht]
  \centerline{\includegraphics[width=16cm]{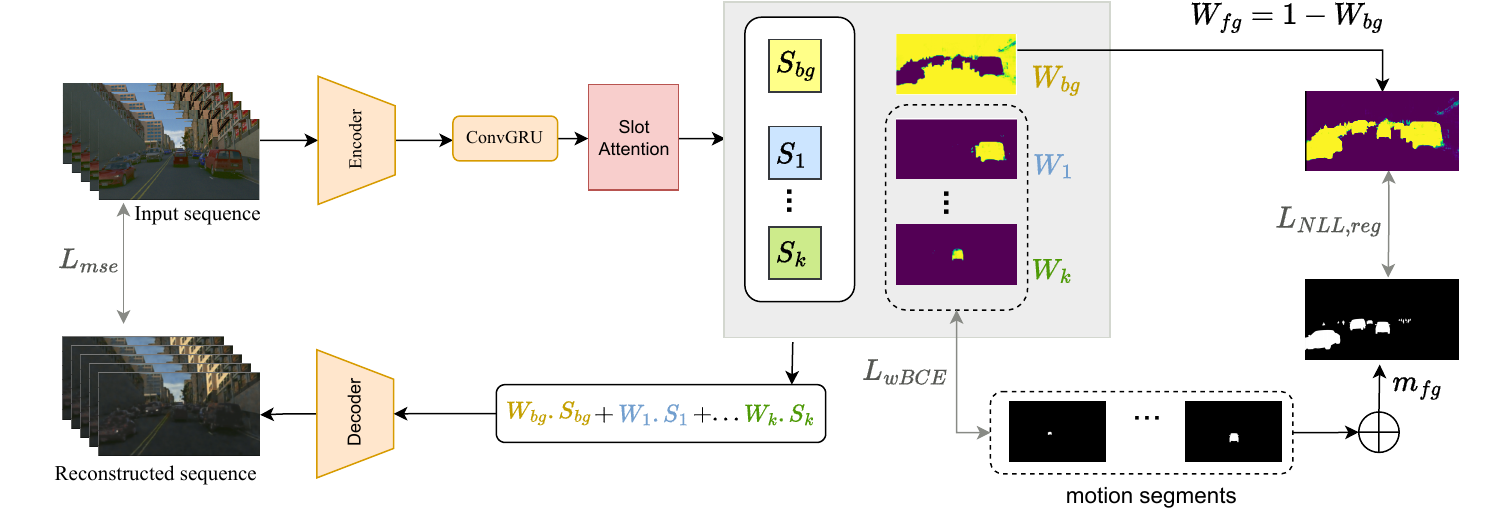}}
  \caption{\textbf{Pipeline of the proposed method.} The input sequence is encoded into a spatio-temporal representation, which is then forwarded to the slot attention module. This produces a set of slots along with their respective attention maps. The separate motion masks individually supervise the attention of \textit{objects} slots, while their sum ($m_{f\!g}$) is generalized to form the \textit{true} foreground, using the $L_{NLL,reg}$ loss. The complement of the learned foreground class is assigned to a specific attention slot $W_{bg}$ so as to isolate the background pattern. Finally, the sum of slots, weighted with their respective attention maps, is decoded to reconstruct the input sequence. B/W masks represent binary supervision masks derived from motion, while masks shown in the viridis colormap are learnable attention maps.}
  \label{fig:global}
\end{figure*}

\noindent Illustrated in figure \ref{fig:global}, our method receives as input $T$ video frames $I^{t}\in \mathbb{R}^{h \times w \times 3}$.
Spatio-temporal representation $H^{t}\in \mathbb{R}^{h'\times w'\times D'}$ for each frame is extracted following the process in \cite{DOM}, with $D'$ the dimension of features output by the convGRU module. These features are then forwarded to the attention module where we propose to jointly learn the object discovery task and the background modeling. The objective is to force background regions to occupy one single slot's attention map, instead of being randomly split across multiple slots. We denote $S_{bg}$ the slot dedicated to the background class and $W_{bg}\in \mathbb{R}^{h'\times w'}$ the corresponding attention map. It is important to note that the motion segments cannot directly provide information on the positions of background regions, since the complement of these masks also contains the static objects we aim to localize (see ablation study in section \ref{ablation_losses}). Instead, these masks can be used as samples of what the object of interest looks like. Therefore, we propose to learn the background class by compelling its complementary mask $W_{f\!g} = 1-W_{bg}$ to contain the \textit{true} foreground class with both moving and static objects. The complementary background mask $W_{bg}$ will thus contain all remaining, non-object regions. On the other hand, the softmax operation applied to the attention maps $W$ ensures their complementarity, preventing the background class from appearing in other \textit{object} slots. Note that $W_{f\!g}$ is an auxiliary attention map only used in the training phase to help the background modeling. It is not involved in the object discovery nor the image reconstruction task.  

 We formulate the foreground modeling as one-class learning problem, since only the positive class is known (some moving objects masks). This paradigm is commonly used for binary classification tasks \cite{OCC}. In our proposed foreground modeling, the samples to classify are the pixels positions in $W_{f\!g}$. The positive class corresponds to pixels in motion (e.g. moving cars), which we propagate to also capture static objects of the same semantic class (e.g. parked cars). For a given frame $I^{t}$, the corresponding moving foreground mask is denoted $m_{f\!g}=\sum_{c=1}^{C}{m_{c}}$, where $C$ is the number of motion masks available for frame $I^{t}$. In order to compel all objects regions to be activated in $W_{f\!g}$ , we use the following negative log likelihood (NLL) loss:
\begin{equation}\label{eq:1}
\begin{aligned}
L_{NLL,reg}(m_{f\!g}, W_{f\!g})&=-\frac{1}{N}\sum_{i=1}^{N} m_{f\!g}(i)\log(W_{f\!g}(i)) \\
&+ \frac{\alpha}{N_{s}}\sum_{j=1}^{N_{s}}{W_{f\!g}(j)}
\end{aligned}
\end{equation}

\noindent where $N=h'\times w'$ is the size of $W_{f\!g}$, $N_{s}$ the number of pixels with no motion information in $m_{f\!g}$ and $\alpha$ a weighting hyper-parameter. The first term in equation \ref{eq:1} is the $NLL$ loss that forces all motion segments to be contained in $W_{f\!g}$, encouraging generalization to visually similar regions (static objects). We can easily predict the collapse that would occur if only this first term is used: the model would converge towards a trivial solution by activating the entire map $W_{f\!g}$, which is not the desired behavior. We rather want the model to only activate objects regions (moving and static) in $W_{f\!g}$. For this, we add as a regularization term in \ref{eq:1} the average activation within the unlabeled regions of $m_{f\!g}$ (i.e. where $m_{f\!g}$ is 0), so as to constraint the model confidence in non-object regions. 
Given a batch size $B$ and $T$ frames per sequence, the final $f\!g/bg$ loss is defined as follows:

\begin{equation}
    L_{f\!g/bg}=\frac{1}{BT}\sum_{b=1}^{B}\sum_{t=1}^{T}{L_{NLL,reg}(m_{f\!g}^{b,t}, W_{f\!g}^{b,t})}
\end{equation}

\subsection{Background-aware motion guided objects discovery} \label{wBCE}
\noindent Similar to \cite{DOM}, the object discovery task is learned through the motion guidance of slots learning: a bipartite matching is performed to associate the motion masks to some of the $K$ \textit{objects} attention maps (excluding the slot assigned to the background). These receive a supervision using the corresponding mask and the Binary Cross-Entropy (BCE) objective function. Different from \cite{DOM}, our method includes a dedicated attention map specifically for the background class, which competes with other slots, while being predominant. This would bias the model towards activating most regions in the background slot, resulting in some objects being lost, specially small ones. To avoid this, we use a weighted $BCE$ loss denoted $L_{wBCE}$, where the weights are set automatically and depend on the object size. Given a motion mask $m$ containing one moving object, which matches the predicted attention map $W$, $L_{wBCE}$ between the two is defined as follows (for the sake of simplification, we denote the \textit{i}-th element of $m$ and $W$ as $m_{i}$ and $W_{i}$ respectively) :

\begin{equation}
    \begin{aligned}
    L_{wBCE}(m, W)&=\frac{1}{N}\sum_{i=1}^{N}(-(2-r).m_{i}\log(W_{i})\\
      &- (1-m_{i})\log(1-W_{i}))
    \end{aligned}
\end{equation}

\noindent where $r$ is the ratio of the number of object pixels to the whole mask size and is computed as $\frac{1}{N}\sum_{i}m_{i}$. In the above, the first loss term is assigned a dynamic weight which depends on the size of the object and varies between $1$ and $2$: the smaller the object in $m$ , the more the model is encouraged  to activate its corresponding pixels in $W$. This weighting has proven effective in maintaining the objects discovery performance, even in the presence of a predominant class (the background) competing with other \textit{object} slots (see ablation study in section \ref{ablation_losses}). \\ 
Finally, the learned slots are broadcasted into 2D maps. The sum of the slots, weighted each by its corresponding attention map, are decoded to reconstruct the input frame. This dense pretext task ensures the activation of all image regions, which further encourages the generalization to non-moving objects. It is learned using a mean squared error (mse) loss between the original and reconstructed video sequence.
The final loss is defined as follows:
\begin{equation}
    L= L_{mse} + L_{wBCE} + L_{f\!g/bg} 
\end{equation}

\section{Experiments}

\noindent We conduct our experiments on two video object discovery benchmarks: ParallelDomain (TRI-PD)\cite{DOM} and KITTI\cite{KITTI}. We further demonstrate the generalizability of the proposed background learning mechanism by integrating it into another state-of-the-art method \cite{MOTOK}. This comparison is conducted under two different settings on TRI-PD, which differ in the source of the motion masks. In the unsupervised setting, referred to as \textit{estimated} in the results tables, these masks are derived from optical flow (see section \ref{Implementation details}). In the second setting denoted \textit{gt}, ground-truth instance masks of moving objects are used as guidance signal. Results in this setting provide an upper-bound for the unsupervised one.
\subsection{Datasets}

\noindent \textbf{ParallelDomain (TRI-PD)}: Introduced by \cite{DOM}, TRI-PD is a recent benchmark for objects discovery in urban driving scenarios. This is a challenging dataset, composed of dense, photo-realistic scenes, which also provides useful support for a variety of visual tasks, as it includes diverse semantic and instance-level annotations. Following\cite{DOM}, we train our object discovery models on a set of 924 video clips, each 200 frames long. Evaluations are performed on a separate test set of 51 video sequences.

\noindent \textbf{KITTI} is a real-world video dataset of urban scenes scenarios, and an active benchmark for various perception tasks. We use for training all raw-data from the KITTI benchmark (without annotations), totalling 147 videos. Following previous works \cite{DOM,MOTOK}, evaluation is conducted on the instance segmentation subset of KITTI, composed of 200 frames.

\subsection{Metrics} \label{metrics}

\noindent \textbf{fg-ARI} and \textbf{all-ARI}: The Adjusted Rand Index (ARI) is a measure used to quantify the similarity between two clusterings (\textit{gt} and \textit{predicted}) in a permutation-invariant way. In the literature of object discovery, studies typically compute the fg-ARI, which stands for ARI in foreground regions. This metric does not account for the segmentation quality in the background regions. We introduce in this paper the computation of the more suitable all-ARI metric, by also incorporating the background class into the ground-truth clusters. In all-ARI, both the foreground objects discovery and the quality of background segmentation are evaluated.

\noindent \textbf{Jaccard score}: The Jaccard score is calculated as the ratio of the size of the intersection to the size of the union of two label sets: the ground truth and the predicted labels. We use this metric to assess the quality of foreground/background classes separation. The final score for each of the two classes is the average Jaccard Score accross all frames.

\subsection{Implementation details} \label{Implementation details}
\noindent \textbf{Base setting (BMOD)}: In this setting, we use a resnet18\cite{resnet} encoder, without pre-training. For a fair comparison, we follow the same training schedule as the method in which we incorporate our background handling mechanism\cite{DOM,MOTOK}. Particularly, we use a batch size of 8 with input sequences of length $T=5$. Frames are resized to $(480\times968)$ and $(368\times1248)$ for TRI-PD and KITTI respectively. The regularization strength $\alpha$, involved in the background modeling, is set to $0.2$ for TRI-PD and $0.4$ in KITTI dataset. In the unsupervised setting, we use the same motion masks as previous methods\cite{DOM,MOTOK}. These are generated using the approach proposed in \cite{SATM} which maps the optical flow to instance masks. The optical flow is computed using RAFT\cite{RAFT} and the mapping is learned on the synthetic dataset FlyingThings3D\cite{FLYING3D}. Following previous methods, models trained on KITTI are initialized with pre-training on TRI-PD dataset, using estimated motion masks. Evaluation on TRI-PD is conducted following the protocol in \cite{DOM} where windows of size $T$ frames (same size as during training) are successively passed to the model. In KITTI dataset, since the test frames are not temporally linked, evaluation is conducted on each frame individually.

\noindent \textbf{Enhanced setting using self-supervised pretraining (BMOD*)}: In this setting, we replace the resnet18 encoder with a ViT-S/14 pretrained using the recent DINOv2 method\cite{dinov2}. For this, we resize the input frames dimensions to adapt to the ViT patch size. Input sizes become $(490\times980)$ and $(378\times1260)$ for TRI-PD and KITTI respectively. We tested integrating the multi-scale features described in DINOv2 paper. Specifically,  we extract features from the last 4 layers of the ViT model, which we concatenate, getting new embedding vectors of size $384\times4$, with spatial dimensions down-sampled with a factor 14. Before passing these features maps to the convGRU, we upscale them to match the spatial dimensions yielded by the resnet18 encoder, resulting in a down-sampling factor of 4.

\subsection{Unsupervised objects discovery}
\begin{figure*}[h]
  \centerline{\includegraphics[width=1\textwidth]{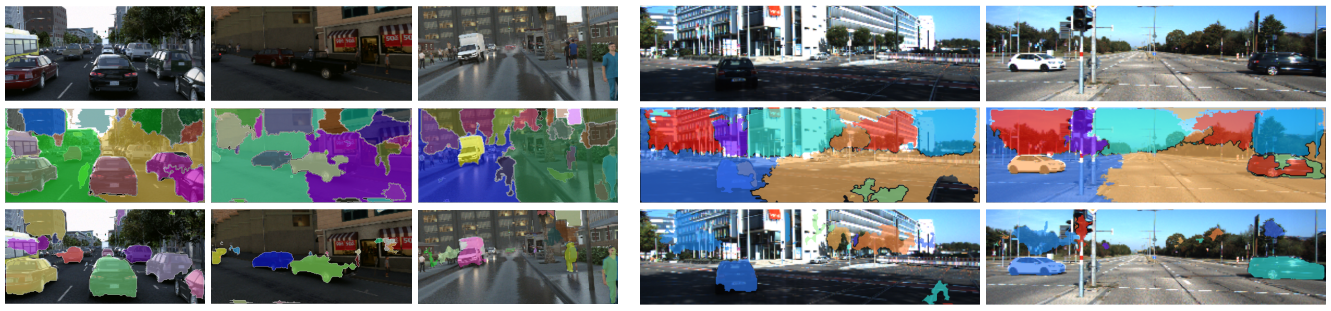}}
  \caption{\textbf{Qualitative comparison under the unsupervised setting on TRI-PD (left) and KITTI dataset (right).} By row: the input frame, results of \cite{DOM} showing both objects and noise segments (indiscernible given the lack of confidence criteria in the unsupervised setting), our segmentation result showing the noise reduction through background modeling. }
  \label{fig:qualitative_results}
\end{figure*}

\begin{table}[h]
\renewcommand{\arraystretch}{1.1} 
  \begin{center}
    {\scalebox{0.75}{
\begin{tabular}{cll}
\toprule
Guidance signal & Method & fg-ARI \\
\hline

- & SlotAttention \cite{SA,DOM} & 10.2  \\
- & MONet \cite{MONET,DOM}       & 11.0  \\
- & SCALOR \cite{SCALOR,DOM}      & 18.6 \\
- & IODINE \cite{IODINE,DOM}      & 9.8  \\
- & MCG \cite{MCG,DOM}     & 25.1  \\

 \cline{1-3}
 \multirow{6}{*}{gt motion} & {Bao et al. \cite{DOM}} & 59.6 \\
 & BMOD (\cite{DOM})  & \underline{74.1}\\
 & BMOD* (\cite{DOM}) & \textbf{83.0} \\
 \cline{2-3}
 & MoTok \cite{MOTOK} & 76.3 \\
 & BMOD (\cite{MOTOK})  & \underline{81.5}\\
 & BMOD* (\cite{MOTOK})  & \textbf{87.5}  \\
 
\midrule

\multirow{6}{*}{estimated motion} & {Bao et al. \cite{DOM}} & 50.9 \\
 & BMOD (\cite{DOM})  & \underline{53.9}\\
 & BMOD* (\cite{DOM})  & \textbf{58.5} \\
 \cline{2-3}
 & MoTok \cite{MOTOK} & \underline{55.1} \\
 & BMOD (\cite{MOTOK})  & 54.2\\
 & BMOD* (\cite{MOTOK})   & \textbf{60.9} \\
 
\bottomrule
\end{tabular}
}}
\end{center}
\caption{Evaluation of objects discovery performance on TRI-PD dataset under two settings. Best results are put in \textbf{bold}, second best \underline{underlined}. BMOD([X]) stands for our method built upon the approach in [X] and  BMOD* is our method enhanced with features from DINOv2\cite{dinov2} pretraining.} \label{FgARI-TRIPD}
\end{table}


\noindent We recall that the primary objective of our work is to enable background handling, for a more precise objects discovery. The results from tables \ref{FgARI-TRIPD} and \ref{FgARI-KITTI} show that this modeling also improves the localization of foreground objects (e.g. we observe $+3\%$ improvement in fg-ARI with BMOD(\cite{DOM}) in the unsupervised setting). This validates our assumption that isolating the background minimizes the presence of random segments in the learned attention maps, which favors a proper learning of the object structure. Although we observe in one test (BMOD(\cite{MOTOK}), \textit{estimated}, table \ref{FgARI-TRIPD}), a slight decrease in fg-ARI ($-0.9$), our approach brings  
a significant gain of performance on the more complete all-ARI metric ($+20.2$, table \ref{FG/BG}).
Our results are further improved when using self-supervised pretraining (\textbf{BMOD*}). This is well-justified given the rich semantic and depth information contained within these features \cite{dinov2}: objects are more easily captured as regions of independent motion, consistent depth, and similar semantics.

\begin{table}[h]
  \begin{center}
    {\scalebox{0.75}{
\begin{tabular}{lcl}
\toprule
Method           & fg-ARI  \\
\midrule
SlotAttention \cite{SA,DOM} & 13.8   \\
MONet \cite{MONET,DOM}       & 14.9   \\
SCALOR \cite{SCALOR,DOM}      & 21.1  \\
IODINE \cite{IODINE,DOM}      & 14.4  \\
MCG \cite{MCG,DOM}      & 40.9   \\
SAVI \cite{SAVI,MOTOK}      & 20.0   \\
SAVI++ \cite{SAVI++,MOTOK}      & 23.9   \\
STEVE \cite{STEVE,MOTOK}      & 11.9   \\
Karazija et al. \cite{karazija}      & 50.8   \\
Karazija et al. (WL) \cite{karazija}      & 51.9   \\
Bao et al \cite{DOM}      & 47.1   \\
BMOD (\cite{DOM})      & \underline{54.7}  \\
BMOD* (\cite{DOM})     & \textbf{60.8}  \\

\bottomrule
\end{tabular}
}}
\end{center}
\caption{Performance comparison of BMOD and previous methods for unsupervised object discovery on KITTI dataset.} \label{FgARI-KITTI}
\end{table}

\subsection{Background modeling using motion cues}
\noindent In this section, we highlight our main contribution, namely learning the object/non-object boundary without human supervision. We use two distinct metrics to evaluate this task, all-ARI and Jaccard Score (see section \ref{metrics}).  
We recall that in our method, the background slot is known since it is constrained by design. Calculation of the Jaccard Score is therefore straightforward. For previous methods, however, no information of the background class is available. For a fair comparison, we consider in these methods as background the largest segment returned in all slots.  Even so, the results in table \ref{FG/BG} show the clear improvement brought by our method in the two tested settings: with motion supervision and unsupervised.
Particularly, incorporating our training mechanism into \cite{DOM} brings considerable all-ARI improvement of $+22.3$  on TRI-PD under the unsupervised setting, and a further enhancement of $+13.5$ on KITTI dataset. 
The other observation we can draw is that, for both fg-ARI and the fg/bg separation tasks, a wide gap remains between the two settings \textit{gt} and \textit{estimated} (the upper-bound results being very high), suggesting strong potential for improvement by addressing the quality of pseudo-labels.

\begin{table}[h]
\renewcommand{\arraystretch}{1.1} 
  \begin{center}
    {\scalebox{0.74}{
\begin{tabular}{lcllll}
\toprule
Dataset & Guidance & Method  & all-ARI & \multicolumn{2}{c}{Jaccard score} \\
& & &  & fg-class & bg-class\\
\hline

\multirow{12}{*}{TRI-PD} & \multirow{6}{*}{gt} & {Bao et al. \cite{DOM}} &  18.1 & 19.3 & 46.2 \\
 & & BMOD (\cite{DOM})  & \underline{79.7} & \underline{73.0} & \underline{95.8}\\
 & & BMOD* (\cite{DOM})  &  \textbf{84.9} & \textbf{78.2} & \textbf{97.6} \\
 \cline{3-6}
 & & MoTok \cite{MOTOK} & 25.2 & 26.5 & 64.3\\
 & & BMOD (\cite{MOTOK})  & \underline{81.7} & \underline{75.7} & \underline{96.5}\\
 & & BMOD* (\cite{MOTOK})  & \textbf{84.0} & \textbf{77.0} & \textbf{97.5}   \\
 
\cline{2-6}

 & \multirow{6}{*}{est} & {Bao et al. \cite{DOM}} & 6.3 & 15.0 & 33.4\\
 & & BMOD (\cite{DOM})   & \underline{28.6} & \textbf{27.2} & \underline{77.5}\\
 & & BMOD* (\cite{DOM})  &  \textbf{29.1} & \underline{26.5} & \textbf{78.7} \\
 \cline{3-6}
 & & MoTok \cite{MOTOK} &  4.7 & 14.8 & 28.5 \\
 & & BMOD (\cite{MOTOK}) &  \underline{24.9} & \underline{25.1} & \underline{73.2}\\
 & & BMOD* (\cite{MOTOK})  &  \textbf{26.7} &  \textbf{25.7} & \textbf{75.2} \\

\midrule
\midrule

\multirow{3}{*}{KITTI} & \multirow{3}{*}{est} & {Bao et al. \cite{DOM}}  & 4.2 & 9.1 & 39.3 \\
 & & BMOD (\cite{DOM}) & \underline{17.8} & \underline{13.7} & \textbf{70.5}\\
 & & BMOD* (\cite{DOM})  &  \textbf{21.7} & \textbf{14.9} & \underline{69.9}\\
 
\bottomrule
\end{tabular}
}}
\end{center}
\caption{Performance comparison of BMOD with previous methods  on  foreground/background separation.} \label{FG/BG}
\end{table}

\section{Ablation and further analysis}
\subsection{Analysis of the composition of objective functions} \label{ablation_losses}

\noindent In this section, we investigate the composition of our objective functions. First, we test a more \textit{naive} way of isolating the background, by explicitly placing the \textbf{non-moving background} in one slot's attention map, using BCE loss. As expected, this method fails to capture most foreground objects, resulting in a critical degradation of fg-ARI. Indeed, the non-moving background in the estimated masks contains all static objects and a few moving but difficult-to-capture instances. All these elements are considered as background in the previous test. Another test is to \textbf{apply regularization to the whole attention map}. One might be motivated to do this to attenuate the noisy regions contained in the estimated motion masks, but this is not optimal as it encourages the model to attenuate activation on object regions too. Finally, we test the variant of our proposed loss functions \textbf{without the dynamically weighted BCE} described in section \ref{wBCE}, by setting a fixed weight of one, instead. As expected, not accounting for object size in our method, when one group/class is predominant (background), encourages the model to place more objects in that group, causing objects to be lost (see table \ref{table:ablation}).\\
all-ARI results are not reported here since they are not informative when there is a significant loss in fg-ARI. In this case, a high all-ARI means that objects have been falsely attributed to the predominant class (background). Our aim, however, is to handle noise in the background without compromising the ability to capture objects.

\begin{table}[h]
  \begin{center}
    {\scalebox{0.75}{
\begin{tabular}{ll}
\toprule
Method & fg-ARI \\
\hline
isolate only non-moving background & 10.1 \\
regularization on the whole map & 48.4 \\
w/o weighted BCE &  45.0\\
Our full approach (BMOD) & \textbf{53.9} \\
\bottomrule
\end{tabular}
}}
\end{center}
\caption{Ablation study on the design of the objective functions on TRI-PD under the unsupervised setting.} \label{table:ablation}
\end{table}

\subsection{Enhancing the unsupervised setting performance: gains from noiseless pseudo-labels}
\noindent In this section we investigate the upper-bound performance that can be achieved in the unsupervised setting, which corresponds to the use of pseudo-masks of moving objects, extracted from optical flow. It is important to note that these estimated labels are subject to a significant amount of noise arising from camera motion. This noise takes the form of random segments which are used to guide the slots learning. In this study, we apply a simple heuristic related to the nature of the analysed scenes, to filter out this noise. Since we're looking to localize objects of a driving scene, which are unlikely to lie at the top of the frame, we apply the heuristic to the position of the objects, filtering out any segment in the first upper tier of the image. As can be seen below, this simple heuristic provides a stronger baseline for all-ARI, while maintaining objects localization performance  (equivalent fg-ARI). This indicates that the method's potential for improvement is related to the quality of the pseudo-labels, justifying further exploration in this area.

\begin{table}[h]
  \begin{center}
    {\scalebox{0.75}{
\begin{tabular}{lllll}
\toprule
Method & fg-ARI & all-ARI & \multicolumn{2}{c}{Jaccard score}  \\
& & & Fg-class & Bg-class \\
\hline
Bao et al.\cite{DOM} & 50.9 & 6.3 & 15.0 & 33.4 \\
BMOD (\cite{DOM}) & \textbf{53.9} & \underline{28.6} & \underline{27.2} & \underline{77.5} \\
BMOD + noiseless pseudo labels & \underline{52.3} & \textbf{58.8} & \textbf{51.5} & \textbf{91.6}\\
\bottomrule
\end{tabular}
}}
\end{center}
\caption{Study of the impact of noise contained in pseudo-labels.} \label{ablate_filter}
\end{table}

\section{Conclusion}
\noindent In this work, we present an objects discovery method that takes into account the particular semantic concept of the background, which is isolated while decomposing the scene into \textit{objects} regions. We showed through the computation of adapted metrics the effectiveness of our method in separating object/non-object regions, without human supervision. In addition, objects localization was found to benefit considerably from the background modeling. This important result is justified by the noise reduction induced by our method, enabling better focusing on object regions. We hope the baseline proposed in this work will motivate further research on background modeling in object discovery. A further analysis showed the potential for scaling the performance of the method by improving the quality of the motion masks. We believe this deserves further exploration in future work. Finally, given the reduced amount of noise among our produced segments, this work opens up the perspective of re-using the discovered objects, for example, with a pseudo-labeling approach.


\section{Acknowledgements}
\noindent This work benefited from the FactoryIA supercomputer financially supported by the Ile-de-France Regional Council

{\small
\bibliographystyle{ieee_fullname}
\bibliography{egbib}
}

\end{document}